\documentclass[journal]{IEEEtran} 
\usepackage{amsmath,amsfonts}
\usepackage{algorithmic}
\usepackage{algorithm}
\usepackage{array}
\usepackage[caption=false,font=normalsize,labelfont=sf,textfont=sf]{subfig}
\usepackage{textcomp}
\usepackage{stfloats}
\usepackage{url}
\usepackage{ulem}
\usepackage{verbatim}
\usepackage{graphicx}
\usepackage{cite}
\usepackage{times}
\usepackage{latexsym}
\usepackage{amsmath}
\usepackage{mathtools}
\usepackage{bbm}
\usepackage{amssymb}
\usepackage{multirow}
\usepackage{float}
\usepackage{enumitem}
\usepackage{bbding}
\usepackage{makecell}
\usepackage{hyperref}
\usepackage{csquotes}
\usepackage{microtype}
\usepackage{inconsolata}

\usepackage[T1]{fontenc}

\begin{document}

\title{Multimodal Mathematical Reasoning with Diverse Solving Perspective}

\author{Wenhao~Shi,
        Zhiqiang~Hu,
        Yi~Bin,
        Guoqing~Wang,
        Xing~Xu\\
        Yang~Yang,~\IEEEmembership{Senior Member,~IEEE}, and
        See-Kiong Ng,~\IEEEmembership{Senior Member,~IEEE}
\thanks{
Wenhao Shi, Yi Bin and Xing Xu are with Tongji University, Shanghai, China. Wenhao Shi is also with National University of Singapore, Singapore. See-Kiong Ng is with the Institute of Data Science, National University of Singapore, Singapore. Zhiqiang Hu is with Tencent, Beijing, China. Guoqing Wang and Yang Yang are with the Center for Future Media, University of Electronic Science and Technology of China, Chengdu, China.
}
\thanks{Yi Bin is the corresponding author. E-mail: yi.bin@hotmail.com}
\thanks{This work has been submitted to the IEEE for possible publication. Copyright may be transferred without notice, after which this version may no longer be accessible.}
}

\markboth{Journal of \LaTeX\ Class Files,~Vol.~28, No.~6, June~2026}%
{Shell \MakeLowercase{\textit{et al.}}: A Sample Article Using IEEEtran.cls for IEEE Journals}


\maketitle

\newcommand{\myet}{\emph{et~al.}}
\newcommand{\myeg}{\emph{e.g.}}
\newcommand{\myie}{\emph{i.e.}}
\newcommand{\myec}{\emph{etc}}

\begin{abstract}
Recent progress in large-scale reinforcement learning (RL) has notably enhanced the reasoning capabilities of large language models (LLMs), especially in mathematical domains. However, current multimodal LLMs (MLLMs) for mathematical reasoning often rely on one-to-one image-text pairs and single-solution supervision, overlooking the diversity of valid reasoning perspectives and internal reflections. In this work, we introduce MathV-DP, a novel dataset that captures multiple diverse solution trajectories for each image-question pair, fostering richer reasoning supervision. We further propose Qwen-VL-DP, a model built upon Qwen-VL, fine-tuned with supervised learning and enhanced via group relative policy optimization (GRPO), a rule-based RL approach that integrates correctness discrimination and diversity-aware reward functions. Our method emphasizes learning from varied reasoning perspectives and distinguishing between correct yet distinct solutions. 
Extensive experiments on the MathVista's minitest and Math-V benchmarks demonstrate that Qwen-VL-DP significantly outperforms prior base MLLMs in both accuracy and generative diversity, highlighting the importance of incorporating diverse perspectives and reflective reasoning in multimodal mathematical reasoning.
\end{abstract}


\section{Introduction}
\IEEEPARstart{L}{arge} language models (LLMs) have demonstrated remarkable abilities in reasoning tasks \cite{wang2023selfconsistency, zhou2023least-to-most, huang2026step}. This has spurred significant interest in their application to solving math problems described in natural language \cite{luo2023wizard, yue2023mammoth, gou2023tora,jiang2023formaltheorem}. Meanwhile, a more challenging direction involves multimodal mathematical reasoning \cite{lu2023mathvista}, where models must interpret various types of images and apply advanced logical skills to address mathematical questions with visual components. Open-source multimodal large language models (MLLMs), such as LLaVA \cite{liu2023llava} and Qwen-VL \cite{bai2023qwenvl}, have achieved strong results on visual question answering benchmarks \cite{guo2023unkvqa}. However, when it comes to intricate mathematical problems that require visual understanding, these models still lag behind close-source counterparts like GPT-4V and Gemini \cite{gpt4v, gemini}.

Humans frequently engage in intuitive chain-of-thought (CoT) processes to address complex reasoning tasks \cite{ericsson1980verbal}. Recent research \cite{jason2022cotreason} has demonstrated that LLMs are capable of exhibiting similar CoT reasoning. By employing straightforward prompting strategies or fine-tuning methods \cite{wang2023selfconsistency,hsieh2023distilling}, CoT can both boost the reasoning abilities of LLMs and increase transparency in their decision-making procedures. Notably, recent progresses, such as OpenAI o1 \cite{OpenAIo1}, have enabled LLMs to generate more elaborate internal CoT sequences.
Despite these successes in natural language contexts, adapting CoT approaches for multimodal tasks remains fully unexplored. In contrast to the rich supply of text-centric CoT data used during language model training, there is a marked shortage of multimodal CoT datasets within predominantly text-based online resources \cite{dai2024instructblip}. This scarcity constrains the development and reasoning capacity of MLLMs.

\begin{figure}[h]
  \centering
\includegraphics[width=1.0\columnwidth]{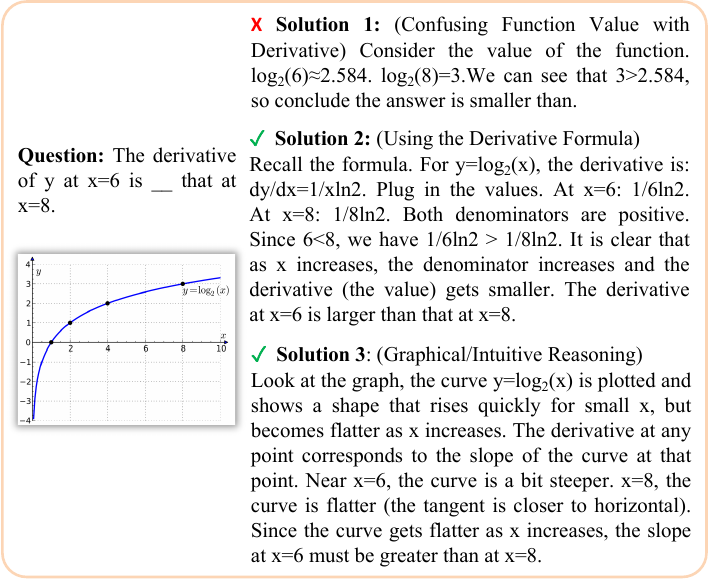}
  \caption{An multimodal mathematical reasoning example with alternative solutions that reaches the final answer. Existing open-source image instruction datasets containing limited solution per image-question,
do not fully exploit diverse solution with reflection to enhance the multimodal mathematical reasoning capabilities of MLLMs.}
  \label{fig:figure1 case}
\end{figure}

Recent advancements in large-scale reinforcement learning (RL) \cite{sutton1998reinforcement} have significantly enhanced the reasoning capacity of LLMs especially within mathematical reasoning tasks. o1~\cite{OpenAIo1} and DeepSeek-R1 \cite{guo2025deepseek} illustrate that extensive RL applied during post-training can lead to substantial gains in complex reasoning performance, in some instances surpassing outcomes achieved via supervised fine-tuning (SFT) \cite{radford2019language}. 
There has been growing interest within the research community to adapt the rule-based RL 
used in DeepSeek-R1
to multimodal scenarios~\cite{chen2025r1v,yang2025r1}. 
These works just explore using final answer and thinking format of image instruction dataset as reward signal.

Furthermore, most existing MLLMs focus on pre-training and post-training by using one-to-one image-text data to improve the final answer accuracy on mathematical reasoning but neglect diverse perspective of internal thought. 
As shown in Figure~\ref{fig:figure1 case}, for an image-question pair, there are usually multiple reasonable inference solutions to reach the final correct answer.
Constrained by limited thinking perspectives tend to derive wrong solution and answer.
Existing open-source image instruction datasets for fine-tuning or reinforcement learning, containing limited solution per image-question,
do not fully exploit diverse solution with reflection to enhance the multimodal mathematical reasoning capabilities of MLLMs.

To bridge the gap, we construct MathV-DP dataset involving a variety of solutions for image-question corresponding to a single thought solution, and train the model Qwen-VL-DP based on the Qwen-VL-7B \cite{Qwen2.5-VL, Qwen2-VL} through supervised fine-tuning and group relative policy optimization (GRPO) \cite{shao2024deepseekmath} as rule-based reinforcement learning.
In addition, the discrimination of diverse  correct solutions and the preference for different correct and incorrect solutions are introduced in the reward function. 
Experiments on MathVista's minitest \cite{lu2023mathvista} and Math-V \cite{wang2024mathv} 
show that learning the correctness, 
diverse skills and reasoning trajectories from multiple solution perspectives significantly improves the accuracy and generation diversity of base MLLMs on multimodal mathematical reasoning.

\section{Related Works}
\subsection{Multimodal Large Language Models}
The rapid progress of LLMs has intensified attention on vision-language integration, with a particular focus on embedding visual understanding within LLMs. The CLIP family~\cite{radford2021clip, li2022blip} played a foundational role by employing contrastive learning across large-scale image-text datasets to align visual and linguistic representations. Subsequent research has emphasized pretraining alignment and the use of visual instruction tuning to enable LLMs to address complex tasks such as visual question answering, art analysis, and multimodal reasoning~\cite{li2024mm,bin2024gallerygpt}. For example, MiniGPT-4~\cite{zhu2023minigpt4} facilitates image-text communication by mapping visual features into the language model space. In parallel, models like LLaVA~\cite{liu2023llava} and Qwen-VL~\cite{bai2023qwenvl, Qwen2-VL} employ learnable projection layers or embed queries to directly interact with visual modalities. 
These strategies capitalize on high-quality datasets for both pretraining and fine-tuning, aiming to improve instruction following and complex visual understanding. Innovations such as mPLUG-Owl~\cite{ye2023mplug}, SPHINX~\cite{lin2023sphinx}, and MiniCPM-V2~\cite{hu2024minicpm} further advance the field by introducing novel forms of grounding data and adopting modular training schemes to reduce hallucinations and strengthen visual grounding. Nevertheless, MLLMs continue to encounter limitations in handling diagram-based mathematical reasoning. To address these challenges, further investigation into the training and quality of image-based instructions is critical for enhancing the reasoning performance of MLLMs.

\subsection{Multimodal Reasoning}
The progress of MLLMs has significantly advanced research in multimodal reasoning~\cite{huang2025vision,wang2025adapting,xu2026mmtot}. A widely adopted strategy involves augmenting existing question-answer datasets in specialized domains to further fine-tune MLLMs. For answer enhancement, rationales have been either human-authored~\cite{zhang2023multicot} or extracted from leading 
LLMs~\cite{wang2024tsciq,chen2023chain,li2024-multimodal-arxiv}. Furthermore, VPD~\cite{hu2023vpd} introduced a method for converting programmatic answer representations into natural language explanations. On the question side, DDCoT~\cite{zheng2023ddcot} employed LLMs to decompose complex queries into simpler sub-questions. Math-LLaVA \cite{mathllava} explored raw visual information presented in images to construct more questions.
To provide a more comprehensive assessment of MLLM multimodal reasoning, several benchmarks have emerged: MathVista~\cite{lu2023mathvista}, and Math-V~\cite{wang2024mathv}
address diverse mathematical reasoning tasks, while MMMU~\cite{yue2023mmmu} spans multiple disciplines. Despite these progresses, open-source MLLMs still exhibit substantial room for improvement in complex multimodal reasoning scenarios.

\subsection{Reinforcement Learning}
Reinforcement learning (RL) \cite{littman1996reinforcement} represents a foundational paradigm within machine learning, wherein an agent interacts with its environment by executing actions, receiving corresponding feedback in the form of rewards, and iteratively updating its policy to optimize cumulative returns over time. 

With the advent of LLMs \cite{brown2020language}, reinforcement learning from human feedback (RLHF) \cite{bai2022training} has emerged as an essential strategy for model fine-tuning, utilizing human-annotated preference data. RLHF commonly incorporates optimization methods like proximal policy optimization (PPO) \cite{schulman2017ppo} and direct preference optimization (DPO) \cite{rafailov2023dpo}, facilitating improved response alignment, coherence, and utility in generated outputs.

Recently, there has been a growing interest in leveraging RL to enhance the reasoning abilities of LLMs \cite{team2025kimi,guo2025deepseek,luo2025ursa, yuan2025more, fan2025sophiavl}, particularly within the scope of mathematical reasoning\cite{sharif2026sight}. The central approach involves designing reward functions or evaluative models that preferentially reinforce high-quality reasoning steps and discourage inadequate reasoning, thereby steering the optimization process toward more organized and comprehensible reasoning patterns through RL techniques. For instance, ReST-MCTS \cite{zhang2024rest} utilizes a process reward model (PRM) to assess the correctness of individual reasoning steps within solution paths. Moreover, recent research indicates that even straightforward rule-based, outcome-level reward functions can serve as robust and informative signals during RL, as demonstrated by DeepSeek-R1 \cite{guo2025deepseek}. DeepSeek-R1 incorporates group relative policy optimization (GRPO) \cite{shao2024deepseekmath} combined with outcome-based reward assessments, effectively advancing the reasoning proficiency of LLMs.
In this work, we focus on further enhancing the reasoning capabilities of MLLMs through reinforcement learning.

\section{Method}
\begin{figure*}[h]
    \centering
    \includegraphics[width=1.025\textwidth]{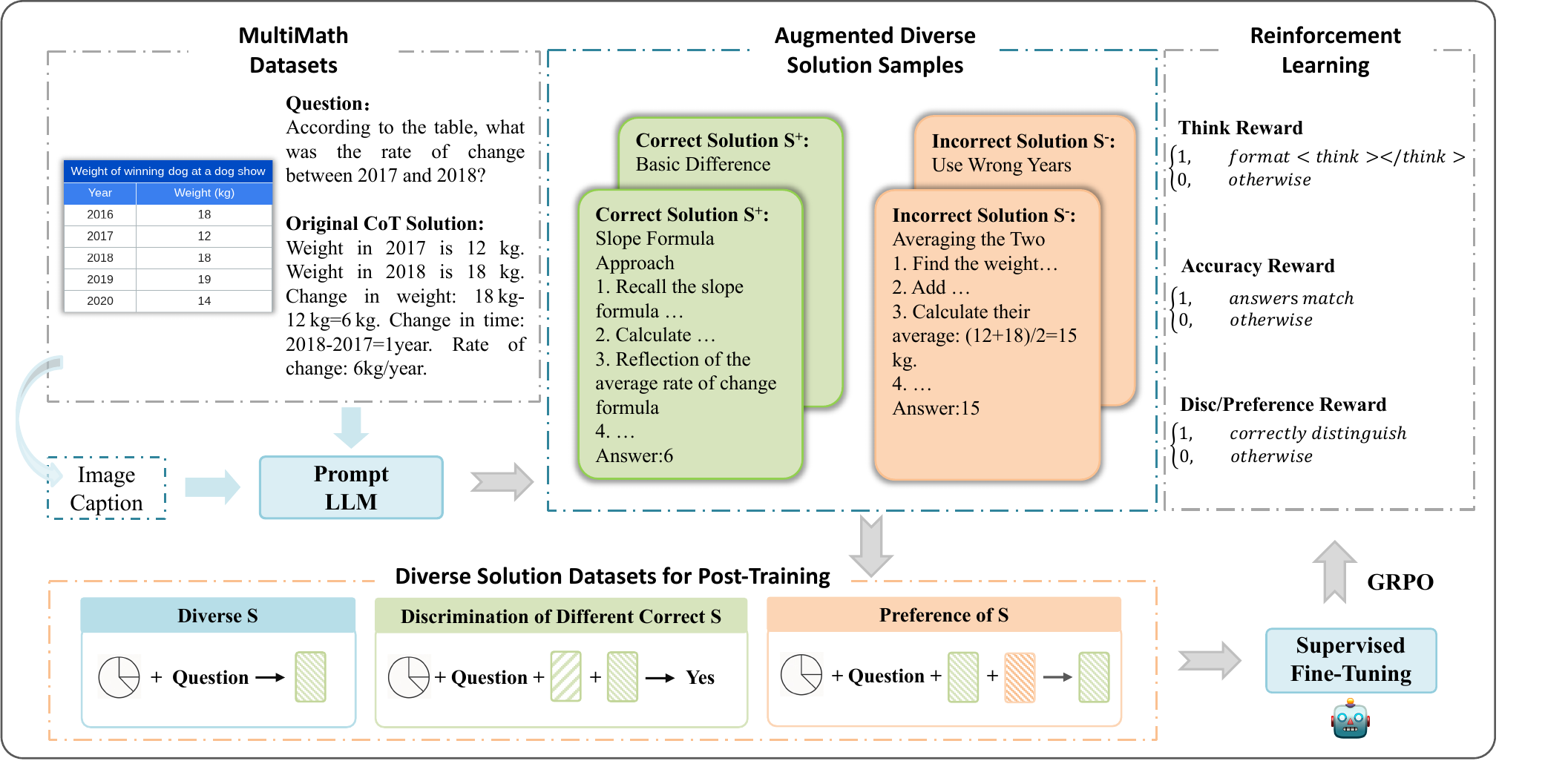}
    \caption{The overall flowchart of the proposed multimodal question-solution data synthesis and post-training. Post-training consists of supervised fine-tuning and rule-based reinforcement learning (GRPO) to learn diverse and reflection reasoning manner.
    }
    \label{fig:framework}
\end{figure*}

Our proposed method is composed of two components: (1) bootstrapping a substantial set of both positive and negative chain-of-thought (CoT) 
solutions with reflection for collected multimodal mathematical question-CoT; and (2) leveraging these new sampled positive solutions, pairs of different positive solutions and pairs of positive-negative solutions to perform post-training on the underlying diverse rationales and to facilitate learning discrimination and preference from identified pairs.
Through the data synthesis and post-training, the MLLM is progressively improved from an initial single solving perspective to a diverse state. The overall framework is depicted in Figure~\ref{fig:framework}.

\subsection{Data Synthesis}
In vision-language reasoning tasks, given an image $I$ and a corresponding question $q$, an MLLM is expected to perform joint reasoning over both modalities to generate a rationale $r$, followed by deriving a final answer $a$. 
However, constructing large-scale datasets comprising high-quality $(I,q,r,a)$ remains a significant challenge, primarily due to the scarcity of well-annotated rationale data. This data bottleneck hinders the post-training enhancement of MLLM reasoning capabilities.
Although MLLMs possess a rudimentary ability for CoT reasoning and self-reflection, leveraging them to generate diverse and high-quality 
$(I,q,r,a)$ samples from existing multimodal mathematical datasets is difficult. Recent advancements in language models, such as DeepSeek-R1, demonstrate strong capabilities in producing coherent, reflective reasoning across extended textual contexts.
Formal languages, characterized by strict syntactic and semantic rules, provide a structured representation that eliminates ambiguity and enforces logical consistency. When visual content is described using formal language, it enables language models to see and reason over image elements more effectively. In our work, we utilize DeepSeek-R1 \cite{guo2025deepseek} to synthesize diverse detailed reasoning chains on samples from the MultiMath-300K dataset \cite{peng2024multimath}. This facilitates the construction of a richer and more diverse set of cross-modal mathematical reasoning samples, culminating in our proposed 40K MathV-DP dataset. The data generation pipeline is illustrated on the left side of Figure~\ref{fig:framework}.

\noindent\textbf{Data Source.} We adopt MultiMath-300K \cite{peng2024multimath} as the primary data source for our data synthesis. This dataset is a large-scale, multimodal, multilingual, multi-level, and multi-step mathematical reasoning benchmark, encompassing a wide range of K-12 level problems. 
It spans nearly the entire K-12 curriculum, covering a broad spectrum of mathematical domains, including arithmetic, algebra, geometry, functions, algorithms, and more.
Compared to existing multimodal mathematics datasets (e.g., Geo170K \cite{gao2023gllava} and MathV360K \cite{mathllava}), the problems in MultiMath-300K are newly curated 
from the real world
and do not overlap with those in previously released datasets
lacking high-quality CoT annotations or containing only final answers.
Each instance is paired with a descriptive image caption to support vision-language alignment, as well as a detailed step-by-step solution.
The availability of formal visual descriptions and CoT annotations in MultiMath-300K with single solution per sample makes it particularly well-suited as seed data for synthesizing diverse solutions from multiple perspectives.
Specifically, we randomly selected 10K samples from them as seed data $\mathcal{D}$.

\begin{figure}[h]
  \centering
\includegraphics[width=0.84\columnwidth]{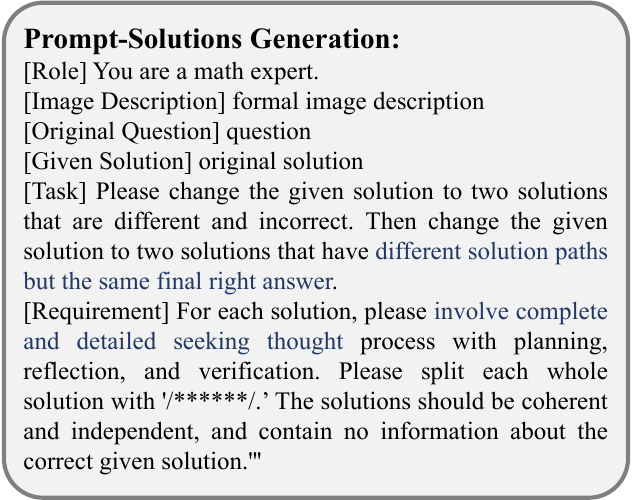}
  \caption{The prompt template used in our DeepSeek-R1
API for generating additional solutions with reflection for each input image description, question and original CoT solution. }
  \label{fig:prompt}
\end{figure}

\noindent\textbf{Diverse Solutions Construction.} Given an image, we
prompt large language reasoning model (i.e., DeepSeek-R1) with its formal dense caption, question and limited original solution to construct more diverse CoT data with reflection. 
The prompt for generating new solutions $s$ is shown in Figure~\ref{fig:prompt}, guiding the model to identify the whole objective and provide a general idea of the plan, propose corrections or alternative reasoning paths, verify consistency between its intermediate reasoning and the final answer. 
Two correct solutions and two incorrect solutions that differ from each other are generated at once for each source sample 
to reduce expenses. 
They are organized into three formats to constitute MathV-DP dataset involving CoT with reflection thinking, discrimination of different correct solutions and preference of solutions.

The correct solution with reflection is first taken out separately with the original image and question. 
The rationale before the final answer in each solution is wrapped with \textit{<think>} and \textit{</think>} tags as $r_{think}$ to form a new set $\mathcal{D}_s^{+}$ totaling 20K: 
\begin{equation}
\begin{aligned}
& \mathcal{D}_s^{+}=\left\{\left(I_i, q_i, r_{think}, a_i\right) \right\}_{i=1}^{|\mathcal{D}|}.
\end{aligned}
\end{equation}
The generated different correct solutions are then concatenated with the instruction $Ins_1$ (i.e., \textit{“Are the solution perspectives of the two solutions dissimilar?”}) to form set $\mathcal{D}_{d}$ totaling 10K used for the calculation of discrimination reward during RL:
\begin{equation}
\begin{aligned}
& \mathcal{D}_{d}=\left\{\left(I_i, q_i, s_{i_1}^{+}, s_{i_2}^{+}, Ins_1, 1\right) \right\}_{i=1}^{|\mathcal{D}|}.
\end{aligned}
\end{equation}
For the data format of correctness preference
and future perference reward calculation, 
one of each of the correct and incorrect solutions is randomly selected and both are concatenated together as a pair in a random back-and-forth order to construct set $\mathcal{D}_{p}$ totaling 10K. Instruction $Ins_2$ is \textit{“Is the former/later solution the correct one?”}:
\begin{equation}
\begin{aligned}
& \mathcal{D}_{p}=\left\{\left(I_i, q_i, s_{i}^{+}, s_{i}^{-}, Ins_2, 1\right) \right\}_{i=1}^{|\mathcal{D}|}.
\end{aligned}
\end{equation}

\subsection{Post-Training}
To improve the multimodal mathematical reasoning capabilities of MLLMs, we propose a two-stage post-training framework comprising supervised fine-tuning followed by rule-based reinforcement learning. In this pipeline, supervised fine-tuning serves to stabilize the model's reasoning ability and learn diverse solving process with reflection, while the subsequent reinforcement learning phase promotes better generalization, preference of solution correctness and diversity in multimodal mathematical reasoning task.

\subsubsection{Supervised Fine-Tuning}
Specifically, we utilize $\mathcal{D}_s^{+}$ with diverse solution perspectives during the supervised fine-tuning stage to guide the model $\mathcal{M}$ toward generating coherent and 
diverse reasoning chains with a negative log-likelihood objective:
\begin{equation}
\mathcal{L}_{\text {SFT}}=-\sum_{(I, q, r, a) \sim \mathcal{D}_s^{+}} \log \mathcal{M}(r, a \mid q, I).
\end{equation}

Supervised fine-tuning not only aligns the model’s outputs with desired formats but also encourages the emergence of more sophisticated multimodal mathematical reasoning reflection behaviors. This establishes a robust foundation for the subsequent RL phase, where rule-based feedback is employed to further refine the model's reasoning abilities.

\subsubsection{Rule-Based Reinforcement Learning}
Building upon the model fine-tuned via 
SFT
we further optimize its structured reasoning capabilities, output validity and diversity of solutions through a rule-based reinforcement learning framework. In particular, we design three reward functions and employ group relative policy optimization (GRPO) \cite{shao2024deepseekmath} for policy updates. 

\noindent\textbf{Accuracy Reward.} The accuracy reward function assesses the correctness of the MLLM’s final output by extracting the predicted answer using regular expressions and comparing it against the ground truth.
We regard multimodal mathematical reasoning as deterministic tasks, the model is required to present the final answer in a predefined format to facilitate consistent and rule-based evaluation. 

\noindent\textbf{Think Format Reward.} To enforce the explicit presence of a reasoning process, the format-based reward function mandates that the MLLM's rationale be encapsulated within predefined delimiters, i.e., \textit{<think>} and \textit{</think>}.
Regularization is used to verify the existence and correct ordering of these markers, thereby ensuring adherence to the required output structure.

\noindent\textbf{Discrimination and Preference Reward.} The discrimination/preference reward function can be viewed as a binary classification task. It is used to evaluate whether the MLLM correctly distinguishes the diversity of different solutions and whether it prefers the correct solution. This reward signal facilitates the model to learn the different perspectives of the solutions and the correctness preference.

\noindent\textbf{Group Relative Policy Optimization.} To ensure stable training with both consistent policy updates and informative reward signals, we adopt group relative policy optimization (GRPO) as our reinforcement learning algorithm.
For each token in the generated sequence, GRPO computes the log-likelihoods under the current policy $\pi(\theta)$ and a reference policy. 
The ratio between these probabilities is then calculated and clipped within the interval $[1-\epsilon, 1+\epsilon]$ to mitigate the risk of overly aggressive updates. 
The reward, normalized to serve as an advantage estimate, is subsequently incorporated into a proximal policy optimization (PPO) objective function:
\begin{equation}
\mathcal{L}_{\text {clip}}=-\mathbb{E}[\min( \text {ratio}_t \cdot  {Ad}_t,  \text {clipratio}_t \cdot  {Ad}_t)],
\end{equation}
where ${Ad}_t$ represents the advantage estimate, quantifying the relative improvement of the chosen action over the expected value under the reference policy.
To further constrain the updated policy from deviating excessively from the reference distribution, a Kullback–Leibler (KL) divergence term is incorporated into the objective, scaled by a coefficient 
$\beta$. The total loss function is defined as:
\begin{equation}
\begin{aligned}
\mathcal{L}_{\mathrm{RL}}(\theta)=-\mathbb{E}[ & \min \left(\operatorname{ratio}_t \cdot Ad_t, \text {clipratio}_t \cdot Ad_t\right)\\
&-\beta \cdot \mathrm{KL}\left(\pi_\theta(y|x), \pi_{\mathrm{ref}}(y|x)\right)].
\end{aligned}
\label{equ_rl_loss}
\end{equation}

GRPO employs a clipping strategy that effectively mitigates drastic changes in the policy, while the incorporation of KL regularization enforces proximity between the updated and reference policies. This dual mechanism enables stable and efficient integration of rule-based rewards, preserving training robustness throughout the optimization process.

\begin{table*}[h]\small
\centering
\caption{Comparison with baselines on the testmini set of MathVista benchmark. Baseline results are
obtained from \cite{lu2023mathvista}. 
The best results in both the close-source and open-source MLLMs are in bold. MathVista is divided in two ways: task type or mathematical skill, and we report the accuracy under each subset.}
\renewcommand{\arraystretch}{1.31}
\setlength{\tabcolsep}{1.8mm}{
\begin{tabular}{c|c|ccccc|ccccccc}
\hline

\multicolumn{1}{c|}{\multirow{2}{*}{\textbf{Model}}} & \multicolumn{13}{c}{\textbf{MathVista}} \\ \cline{2-14}
\multicolumn{1}{c|}{} & ALL& FQA& GPS& MWP& TQA& VQA& ALG &ARI &GEO &LOG &NUM &SCI &STA         \\ \hline

\multicolumn{14}{c}{\multirow{1}{*}{\textit{Heuristics Baselines}}} \\ \hline
\multicolumn{1}{c|}{Random Chance} & 17.9& 18.2& 21.6& 3.8& 19.6& 26.3& 21.7 &14.7 &20.1 &13.5 &8.3 &17.2 &16.3         \\
\multicolumn{1}{c|}{Frequent Guess \cite{lu2023mathvista}} & 26.3& 22.7& 34.1& 20.4& 31.0& 24.6& 33.1 &18.7 &31.4 &24.3 &19.4 &32.0 &20.9         \\
\multicolumn{1}{c|}{Human} & 60.3&	59.7&	48.4&	73.0&	63.2&	55.9&	50.9	&59.2&	51.4&	40.7&	53.8&	64.9&	63.9         \\ \hline

\multicolumn{14}{c}{\multirow{1}{*}{\textit{Close-Source Multimodal Large Language Models (MLLMs)}}} \\ \hline
\multicolumn{1}{c|}{Gemini 1.0 Nano 2 \cite{team2023gemini}} & 30.6&	28.6&	23.6&	30.6&	41.8&	31.8&	27.1&	29.8&	26.8&	10.8&	20.8&	40.2&	33.5      \\
\multicolumn{1}{c|}{Qwen-VL-Plus \cite{bai2023qwenvl}} &43.3 &	\textbf{54.6} &	38.5&	31.2&	55.1 & 34.1 & 39.1&	32.0&	39.3&	18.9 &	26.4&	59.0 &	56.1\\ 
\multicolumn{1}{c|}{Gemini 1.0 Pro \cite{team2023gemini}} & 45.2&	47.6&	40.4&	39.2&	61.4&	\textbf{39.1}&	45.2&	38.8&	41.0&	10.8&	\textbf{32.6}&	54.9&	\textbf{56.8}     \\
\multicolumn{1}{c|}{Claude 3 Haiku \cite{anthropic2024claude}} & 46.4&	-&	-&	-&	-&	-&	-&	-&	-&	-&	-&	-&	-     \\
\multicolumn{1}{c|}{GPT-4V \cite{gpt4v}} & 49.9 &	43.1&	\textbf{50.5}&	\textbf{57.5}&	\textbf{65.2}&	38.0&	\textbf{53.0}&	\textbf{49.0}&	\textbf{51.0}&	\textbf{21.6}&	20.1&	\textbf{63.1}&	55.8    \\
\multicolumn{1}{c|}{GPT-4o \cite{chatgpt}} & 63.8&	-&	-&	-&	-&	-&	-&	-&	-&	-&	-&	-&	-    \\
\multicolumn{1}{c|}{Llama 4 Maverick  \cite{llama4}} & 73.7&	-&	-&	-&	-&	-&	-&	-&	-&	-&	-&	-&	-    \\
\multicolumn{1}{c|}{OpenAI o1 \cite{OpenAIo1}} & \textbf{73.9}&	-&	-&	-&	-&	-&	-&	-&	-&	-&	-&	-&	-    \\
 \hline

\multicolumn{14}{c}{\multirow{1}{*}{\textit{Open-Source Multimodal Large Language Models (MLLMs)}}} \\ \hline
\multicolumn{1}{c|}{IDEFICS-Instruct \cite{laurenccon2024obelics}} & 19.8& 21.6& 21.1& 6.5& 25.9& 24.0& 22.1 &15.0 &19.8 &18.9 &9.9 &24.6 &18.1         \\
\multicolumn{1}{c|}{mPLUG-Owl-7B \cite{ye2023mplug}} & 22.2& 22.7& 23.6& 10.2& 27.2& 27.9& 23.6 &19.2 &23.9 &13.5 &12.7 &26.3 &21.4 \\
\multicolumn{1}{c|}{miniGPT4-7B \cite{zhu2023minigpt4}} & 23.1&	18.6&	26.0& 13.4&	30.4&	30.2&	28.1&	21.0&	24.7&	16.2&	16.7&	25.4&	17.9 \\
\multicolumn{1}{c|}{LLaMA-Adapter-V2 \cite{gao2023llama}} & 23.9&	21.2&	25.5&	11.3&	32.3&	31.8&	26.3&	20.4&	24.3&	24.3&	13.9&	29.5&	18.3 \\
\multicolumn{1}{c|}{LLaVAR-13B \cite{zhang2023llavar}} & 25.2&	21.9&	25.0&	16.7&	34.8&	30.7&	24.2&	22.1&	23.0&	13.5&	15.3&	42.6&	21.9 \\
\multicolumn{1}{c|}{InstructBLIP-7B \cite{dai2024instructblip}} & 25.3&	23.1&	20.7&	18.3&	32.3&	35.2&	21.8&	27.1&	20.7&	18.9&	20.4&	33.0&	23.1 \\
\multicolumn{1}{c|}{LLaVA-13B \cite{liu2023llava}} & 26.1&	26.8&	29.3&	16.1&	32.3&	26.3&	27.3&	20.1&	28.8&	24.3&	18.3&	37.3&	25.1 \\
\multicolumn{1}{c|}{SPHINX-V1-13B \cite{lin2023sphinx}} & 27.5&	23.4&	23.1&	21.5&	39.9&	34.1&	25.6&	28.1&	23.4&	16.2&	17.4&	40.2&	23.6 \\
\multicolumn{1}{c|}{LLaVA-1.5-13B \cite{liu2024llava-1.5}} & 27.7&	23.8&	22.7&	18.3&	40.5&	30.2&	25.3&	26.4&	22.8&	21.6&	26.4&	35.3&	23.6 \\
\multicolumn{1}{c|}{OmniLMM-12B \cite{omnilmm}} & 34.9&	45.0&	17.8&	26.9&	44.9&	39.1&	23.1&	32.3&	20.9&	18.9&	27.8&	45.9&	44.2 \\
\multicolumn{1}{c|}{SPHINX-V2-13B \cite{lin2023sphinx}} &36.7&	54.6 &	16.4&	23.1&	41.8&	43.0 &	20.6&	33.4&	17.6&	24.3&	21.5&	43.4&	51.5\\
\multicolumn{1}{c|}{MiniCPM-V2 \cite{hu2024minicpm}} &40.6&	53.2&	26.0&	37.1&	44.3&	39.1&	28.5&	33.1&	28.0&	10.8&	\textbf{39.6}&	\textbf{48.4}&	\textbf{51.8} \\

\multicolumn{1}{c|}{G-LLaVA-13B \cite{gao2023gllava}} &- &	- &	56.7&	-&	-& - & -&	-&	-&	- &	-&	- &	-\\ 
\multicolumn{1}{c|}{LLaVA-Next-34B \cite{liu2024llavanext}} & 46.5&	-&	-&	-&	-&	-&	-&	-&	-&	-&	-&	-&	- \\ 
\multicolumn{1}{c|}{Math-LLaVA \cite{mathllava}} & 46.6	& 37.2	& 57.7 &	56.5	& 51.3	&33.5 &	53	& 40.2 &	56.5 &	16.2 &	33.3 &	49.2 &	43.9 \\
\multicolumn{1}{c|}{Math-PUMA-7B \cite{mathpuma}} & 47.9	& -	& 48.1&	-	& -	&- &	-	& - &	47.3 &	- &	- &	- &	- \\
\multicolumn{1}{c|}{ Multimath-7B \cite{peng2024multimath}} & 50.0	& -	& 66.8&	61.8	& -	&- &	-	& - &	- &	- &	- &	- &	- \\
\multicolumn{1}{c|}{Mulberry-7B \cite{yao2024mulberry}} & 63.1	& -	& -&	-	& -	&- &	-	& - &	- &	- &	- &	- &	- \\
\multicolumn{1}{c|}{Qwen2-VL-7B \cite{Qwen2-VL}} & 57.6	& 65.1	& 41.8 &	66.1	& 60.1	&53.7 &	44.5	& 56.4 &	43.1 &	24.3 &	39.6 &	63.1 &	69.4 \\
\multicolumn{1}{c|}{Qwen2.5-VL-7B \cite{Qwen2.5-VL}} & 68.2	& 72.5	& 66.8 &	76.9	& 66.7	&54.3 &	70.1	&68.7  &66.9	 &26.9	 &	43.0 &65.7	 &76.1	 \\
\hline
\multicolumn{1}{c|}{LLaVA-1.5-DP} & 42.2&	32.3&	56.2&	51.6&	45.6&	39.7&	43.8&	41.6&	46.0&	15.4&	38.9&	46.7&	40.9 \\
\multicolumn{1}{c|}{Qwen2-VL-DP} & 60.9	& 70.7	&56.4  &69.8		&64.6 	&48.7 &50.9		&60.4  &47.2	 &25.4	 &40.3	 &65.6	 &71.1	 \\
\multicolumn{1}{c|}{\textbf{Qwen2.5-VL-DP}} & \textbf{70.4}	& \textbf{72.8}	& \textbf{72.6} &	\textbf{77.2}	& 	\textbf{68.5}&\textbf{54.5} &	\textbf{71.1}	& \textbf{69.6} &\textbf{69.3} &	\textbf{27.0} &\textbf{43.1}	 &\textbf{66.9}	 &\textbf{77.2}	 	 \\
\hline

\end{tabular}}

\vspace{-10pt}
\label{tab:baseline}
\end{table*}

\begin{table*}[h]\small
\centering
\caption{Performance Comparison on the Math-V benchmark with the accuracy metric across various mathmatical subjects. Baseline results are obtained from \cite{wang2024mathv}.
The best results in both the close-source and open-source MLLMs are in bold.}

\renewcommand{\arraystretch}{1.55}
\setlength{\tabcolsep}{0.80mm}{
\begin{tabular}{c|c|cccccccccccccccc}
\hline

\multicolumn{1}{c|}{\multirow{2}{*}{\textbf{Model}}} & \multicolumn{16}{c}{\textbf{Math-V}} \\ \cline{2-18}
\multicolumn{1}{c|}{} &ALL &Alg &AnaG& Ari& CG& Comb& Cnt& DG& GT& Log& Angle& Area& Len& SG &Sta& Topo& TG         \\ \hline
\multicolumn{18}{c}{\multirow{1}{*}{\textit{Heuristics Baselines}}} \\ \hline
\multicolumn{1}{c|}{Random Chance} &7.2& 1.5& 11.9& 7.1& 9.7 &4.8& 6.0& 22.1& 1.1& 7.6& 0.6& 9.4& 6.7& 8.2 &8.6& 13.0& 7.1    \\
\multicolumn{1}{c|}{Human} & 68.8 &	55.1&	78.6&	99.6&	98.4&	43.5&	98.5&	91.3&	62.2&	61.3&	33.5&	47.2&	73.5&	87.3&	93.1&	99.8&	69.0         \\ \hline

\multicolumn{18}{c}{\multirow{1}{*}{\textit{Close-Source Multimodal Large Langugae Models (MLLMs)}}} \\ \hline

\multicolumn{1}{c|}{Qwen-VL-Plus \cite{bai2023qwenvl}} &10.7&	11.3&	17.9&	14.3&	12.7&	4.8&	10.5&	15.4&	8.9&	14.3&	11.6&	6.4&	10.0&	14.3&	6.9&	8.7&	11.3	\\ 
\multicolumn{1}{c|}{Qwen-VL-Max \cite{bai2023qwenvl}} &15.6&	10.7&	19.1&	20.0&	16.9&	12.5&	17.9&	16.4&	12.2&	21.0&	13.3&	14.2&	19.8&	11.5&	20.7&	13.0&	17.3	\\ 
\multicolumn{1}{c|}{Gemini Pro \cite{team2023gemini}} & 17.7&	15.1&	10.7&	20.7&	20.1&	11.9&	7.5&	20.2&	21.1&	16.8&	19.1&	19.0&	20.0&	14.3&	13.8&	17.4&	20.8     \\

\multicolumn{1}{c|}{GPT-4V \cite{gpt4v}} & 22.8 &	27.3 &	32.1&	35.7&	21.1&	16.7&	13.4&	22.1&	14.4&	16.8&	\textbf{22.0} &	22.2&	20.9&	23.8 &	24.1 &	21.7 &	\textbf{25.6}    \\

\multicolumn{1}{c|}{GPT-4o \cite{chatgpt}} &\textbf{30.4}& \textbf{42.0}&	\textbf{39.3}&	\textbf{49.3}&	\textbf{28.9}&	\textbf{25.6}&	\textbf{22.4}&	\textbf{24.0}&	\textbf{23.3}&	\textbf{29.4}&	17.3&	\textbf{29.8}&	\textbf{30.1}&	\textbf{29.1}&	\textbf{44.8}&	\textbf{34.8}&	17.9    \\

 \hline

\multicolumn{18}{c}{\multirow{1}{*}{\textit{Open-Source Multimodal Large Langugae Models (MLLMs)}}} \\ \hline

\multicolumn{1}{c|}{SPHINX-V2-13B \cite{lin2023sphinx}} &9.7&	6.7&7.1&	12.9&	7.5&	7.7&	6.0&	9.6&	\textbf{16.7}&	10.1&	11.0&	11.8&	12.5&	8.2&	8.6&	8.7&	6.0\\

\multicolumn{1}{c|}{LLaVA-1.5-13B \cite{liu2024llava-1.5}} & 11.1&	7.0&	14.3&	14.3&	9.1&	6.6&	6.0&	13.5&	5.6&	13.5&	10.4&	12.6&	14.7&	11.5&	13.8&	13.0&	10.7\\

\multicolumn{1}{c|}{Math-LLaVA \cite{mathllava}} & 15.7&	9.0&	20.2&	15.7&	18.2&	10.1&	10.5&	16.4&	14.4&	16.0&	20.2&	18.4&	17.6&	9.4&	24.1&	\textbf{21.7}&	17.9   \\ 

\multicolumn{1}{c|}{Qwen2-VL-7B \cite{Qwen2-VL}} & 16.3& 11.3&	24.9&	15.7&	16.9&	10.1&	11.9&	16.4&	15.6&	19.3&	22.5&	16.4&	22.5&	14.3&	17.2&	4.4&	20.8\\

\multicolumn{1}{c|}{Qwen2.5-VL-7B \cite{Qwen2.5-VL}} & 25.0&	22.0&	29.8&	32.1&	19.5&	18.5&	16.4&	22.1&	11.1&	25.2&	\textbf{29.3}&	27.6&	28.5&	22.9&	34.5&	17.4&	22.0\\

\multicolumn{1}{c|}{Qwen2-VL-DP} 
& 17.7&	15.2&	20.8&	20.8&	20.2&	12.0&	7.9&	20.3&	21.2&	16.9&	19.2&	19.1&	23.2&	14.4&	13.9&	17.5&	20.9\\

\multicolumn{1}{c|}{\textbf{Qwen2.5-VL-DP}} & \textbf{26.9}&	\textbf{23.3}&	\textbf{30.8}&	\textbf{32.2}&	\textbf{20.6}&	\textbf{27.3}&	\textbf{17.4}&	\textbf{23.9}&	\textbf{22.9}&	\textbf{28.6}&	28.9&	\textbf{30.9}&	\textbf{28.8}&	\textbf{28.7}&	\textbf{37.9}&	18.4&	\textbf{23.2}\\

\hline
\end{tabular}}

\vspace{-10pt}
\label{tab:math-v}
\end{table*}

\section{Experiments}
\subsection{Model and Implementation}
We utilize the Qwen2-VL \cite{Qwen2-VL} and Qwen2.5-VL \cite{Qwen2.5-VL} series as our baseline architectures and focus our evaluation on the 7B parameter scale to assess the effectiveness of our proposed method. 
Both the projection layer and the language model parameters are trainable. Supervised fine-tuning stage is performed with a batch size of 16, a learning rate of 
2e-5 over 1 epoch. 
During the reinforcement learning stage, we generate 4 rollouts per query with a sampling temperature of 1.0. The maximum sequence length is set to 1024 to ensure the model has sufficient capacity to produce complete reasoning solution. Both the policy and reference models are initialized from the same base model, with the reference model held frozen during RL training. The policy model is fine-tuned using a learning rate of 1e-6 and a batch size of 4. The KL divergence regularization coefficient $\beta$ in Eq.~\ref{equ_rl_loss} is set to 0.04 by default. All experiments are conducted on NVIDIA H100 GPU with 80GB of memory.




\subsection{Evaluation and Metrics}
We assess our model's performance in a zero-shot setting 
similar to other models
using the minitest subset of the MathVista benchmark \cite{lu2023mathvista}. This subset comprises 1,000 items, including 540 multiple-choice problems and 460 free-response questions requiring answers in the form of integers, floating-point numbers, or lists.
MathVista is designed to comprehensively evaluate the multimodal mathematical capabilities of MLLMs, covering diverse reasoning categories such as algebraic (ALG), arithmetic (ARI), geometric (GEO), logical (LOG), numeric commonsense (NUM), scientific (SCI), and statistical reasoning (STA).
Additionally, its questions are distributed across various subtypes, including Figure Question Answering (FQA), Geometry Problem Solving (GPS), Math Word Problem (MWP), Textbook Question Answering (TQA), and Visual Question Answering (VQA). 
For evaluation, we leverage GPT-4 \cite{chatgpt} to extract final answers or selected choices from model responses in a few-shot manner \cite{lu2023mathvista} and compute accuracy by verifying the correspondence between predicted and grounded answers.
In addition, we perform evaluations on Math-V \cite{wang2024mathv}.
Math-V contains 3,040 visual-context math problems curated from authentic math competitions. 

Accuracy evaluation mainly depends on the final answer of the MLLM output, we also use the \textit{effective semantic diversity} metric \cite{shypula2025evaluating} to assess the diversity of the MLLM's output solutions.
For each input, model generates $K$ responses $G_i=\left\{g_i^1, g_i^2, \ldots, g_i^K\right\}$.
We then adopt the following pairwise diversity score:
\begin{equation}
\operatorname{Div}_{\text {pair }}\left(G_i\right)=\frac{1}{\binom{K}{2}}\sum_{\substack{1 \leq j<k \leq K}} d_{\text {sem}}\left(g_i^j, g_i^k\right),
\end{equation}
where $d_{\text {sem}}$ is semantic distance function. It is obtained by Sentence Transformer \cite{reimers-2019-sentence-bert}, which is 1 if semantically dissimilar and 0 otherwise.
This pairwise evaluation strategy incorporates normalization over the total number of candidate pairs, thereby ensuring robustness against fluctuations in the number of valid outputs generated for different prompts.
The overall diversity of a model on the benchmark is then computed by averaging all pairwise diversity scores.


\section{Results and Analysis}
\subsection{Main Comparison on Accuracy}
We compare Qwen-VL-DP with other MLLMs on the minitest split of the MathVista benchmark in Table~\ref{tab:baseline}. 
As shown in the table, open-source MLLMs such as instructBLIP \cite{dai2024instructblip} and LLaVA-1.5 \cite{liu2023llava} have poor performance in multimodal mathematics, with overall accuracy lower than 30\%.
Compared to the base model, Qwen2.5-VL-7B, with superior multimodal mathematical ability, Qwen2.5-VL-DP achieves 70.4\% overall accuracy with a improvement of 2.2\%. 
LLaVA-1.5-DP also obtains improvement of 14.5\% compared with base model LLaVA-1.5-13B.
We also conducted 10 independent inference runs on Qwen2.5-VL-7B and Qwen2.5-VL-DP, observing an average improvement of 2.5\% (±0.9\%). The 95\% confidence interval for the performance gain is (1.94\%, 3.06\%).
More surprisingly, the proposed Qwen2.5-VL-DP model outperforms close-source models GPT-4V and GPT-4o \cite{gpt4v}, even achieving comparable performance to OpenAI o1  \cite{OpenAIo1}, the most powerful close-source MLLMs with the ability of detailed thinking.
The results on Math-V are shown in Table~\ref{tab:math-v}.
Qwen2.5-VL-DP demonstrates substantial performance gains over its base model, narrowing the gap with state-of-the-art models such as GPT-4V and GPT-4o.
The excellent performance of Qwen-VL-DP indicates that the high-quality data synthesis of solutions with diverse perspective 
is effective in improving MLLM's multimodal mathematical reasoning capabilities. 

\subsection{Comparision on Generation Diversity}
The proposed Qwen-VL-DP model demonstrates exceptional performance on multimodal mathematical reasoning tasks. To further evaluate its ability to generate diverse reasoning processes, we conduct experiments using the effective semantic diversity metric on the MathVista minitest subset. For each input sample, we generate $K=3$, $5$, and $10$ responses and compute the corresponding pairwise diversity score for final averaging.
Table~\ref{tab:semantics diver} compares the effective semantic diversity of the Qwen-VL base model, the supervised fine-tuned model, the model trained with GRPO only, and the final Qwen-VL-DP model obtained through our two-stage post-training framework using MathV-DP. The results show that both supervised fine-tuning and reinforcement learning with solution data containing diverse reasoning perspectives can improve the generative diversity of the base model. Moreover, the full Qwen-VL-DP model consistently achieves the highest diversity scores while simultaneously maintaining superior reasoning accuracy.
These findings suggest that the proposed MathV-DP dataset and post-training framework effectively enhance both reasoning quality and response diversity. We attribute this improvement to two factors. First, SFT enables the model to learn multiple valid solution perspectives from the diverse reasoning trajectories in MathV-DP. Second, the subsequent rule-based reinforcement learning stage further encourages the model to distinguish between solutions of different trajectories and to prefer more reliable reasoning paths through discrimination and preference rewards. As a result, Qwen-VL-DP is able to generate more diverse yet accurate reasoning processes.

\begin{table}[h]\small
\centering
\caption{Effective semantic diversity scores for Qwen-VL models evaluated in our experiments.}
\renewcommand{\arraystretch}{1.3}
\setlength{\tabcolsep}{1.15mm}{
\begin{tabular}{c|c|c|c}
\hline

\multicolumn{1}{c|}{\multirow{1}{*}{\textbf{Model}}} &\textbf{Diver@3}&\textbf{Diver@5} &\multicolumn{1}{c}{\textbf{Diver@10}}   \\ \hline 
\multicolumn{1}{c|}{Qwen2-VL-7B} & 27.64 & 30.18 & 31.33   \\
\multicolumn{1}{c|}{Qwen2-VL-SFT} & 33.72 & 35.63 & 35.75   \\
\multicolumn{1}{c|}{Qwen2-VL-GRPO} & 35.05  & 36.08 & 37.11  \\ 
\multicolumn{1}{c|}{Qwen2-VL-DP} &  37.48 & 38.97 & 39.16 \\
\hline
\multicolumn{1}{c|}{Qwen2.5-VL-7B} & 33.29 & 34.76 & 36.89   \\
\multicolumn{1}{c|}{Qwen2.5-VL-SFT} & 39.46 &  39.78 & 39.81   \\
\multicolumn{1}{c|}{Qwen2.5-VL-GRPO} & 39.02  & 39.49  & 39.73 \\ 
\multicolumn{1}{c|}{Qwen2.5-VL-DP} & \textbf{40.42}  &  \textbf{41.44} & \textbf{41.58} \\

\hline
\end{tabular}}

\vspace{-5pt}
\label{tab:semantics diver}
\end{table}

\subsection{Effectiveness of Disc./Preference Reward}
To assess the individual impacts of the discrimination and preference rewards in GRPO training, we conduct ablation studies on Qwen2.5-VL-7B by selectively removing each reward component. Specifically, we evaluate three variants: (1) removing both discrimination and preference rewards, (2) retaining only the discrimination reward, and (3) retaining only the preference reward. As reported in Table~\ref{tab:abre}, all three variants achieve lower accuracies on both MathVista and Math-V compared with the full Qwen-VL-DP model. These results demonstrate that each reward component contributes positively to performance, while their combination leads to better reasoning capabilities.

The discrimination reward encourages the model to identify and explore diverse yet valid solution trajectories, thereby improving reasoning diversity and exploration. In contrast, the preference reward guides the model to assign higher confidence to correct solutions over incorrect ones, enhancing solution quality and reliability. When combined, the two rewards jointly promote both diversity and correctness in the reasoning process. The consistent performance degradation observed after removing either component further validates their complementary effects and highlights the importance of explicitly rewarding both aspects during reinforcement learning.
\begin{table}[h]\small
\centering
\caption{Performance comparison by isolating discrimination and preference rewards.}
\renewcommand{\arraystretch}{1.3}
\setlength{\tabcolsep}{1.5mm}{
\begin{tabular}{c|c|c|c|c}
\hline
			
\multicolumn{1}{c|}{\multirow{1}{*}{\textbf{Reward}}} & w/ Both & w/o Disc. & w/o Pref. & w/o Both \\ \hline
\multicolumn{1}{c|}{\textbf{MathVista}} & 70.4 & 70.1 & 70.2 & 69.9 \\ 
\multicolumn{1}{c|}{\textbf{Math-V}} & 26.9 & 25.9 & 26.1 & 25.2 \\ \hline
\end{tabular}}

\label{tab:abre}
\end{table}

\subsection{Enhancements from SFT and RL}
\begin{figure}[]
    \centering
   \includegraphics[width=1.0\columnwidth]{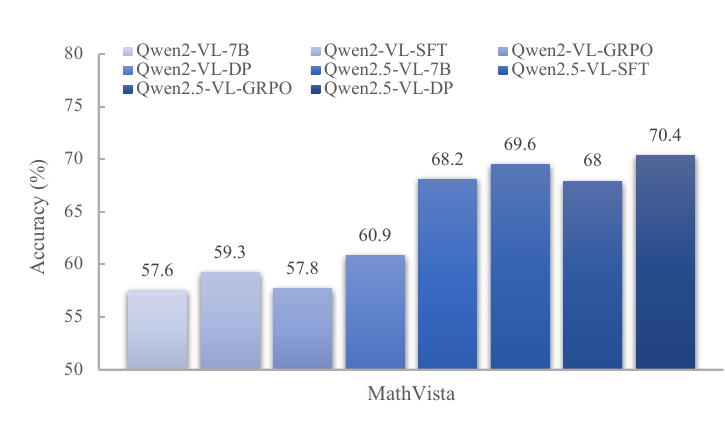}
    \caption{Accuracy of Qwen-VL model adopting different post-training strategies on MathVista.
    }
    \vspace{-10pt}
    \label{fig:acc}
\end{figure}

We conduct ablation study across three training paradigms: (1) supervised fine-tuning (SFT) on our curated dataset, (2) SFT followed by GRPO, and (3) RL applied in isolation. As shown in Figure~\ref{fig:acc}, MLLM by SFT 
demonstrates improvements on the MathVista. 
Applying RL to the SFT model yields further gains, suggesting that RL facilitates deeper and more varied deductive reasoning.
These progressive enhancements underscore the complementary strengths of SFT and RL: while SFT provides a stable foundation by aligning the model with diverse high-quality reasoning perspectives, RL further strengthens these abilities by promoting advanced cognitive behaviors. In contrast, applying RL without prior SFT leads to suboptimal performance, likely due to the absence of a structured reasoning baseline. Overall, integrating SFT with RL emerges as an effective paradigm for enhancing the MLLM’s mathematical reasoning ability.

\subsection{Evaluation on Generated Dataset}
Human evaluation is a widely adopted method for assessing the quality of synthesized data \cite{long2024llms}. We conduct a manual review of 1,000 randomly selected samples by five annotators to ensure the objectivity of checks. Like students marking exam papers, our evaluation emphasizes key aspects including overall correctness of solution and distinctions between correct and incorrect outputs. The average scores across these dimensions were 0.82 (±0.8\%), 0.97 (±0.4\%), and 0.95 (±0.5\%) (on a scale of 0, 1), indicating that the generated solutions are generally of high quality and it is sufficiently reliable to enrich the solution space by introducing greater diversity. The synthesized solutions demonstrate varied mathematical reasoning approaches, offering a broader set of reasoning patterns that enhance the base model’s capabilities.
Additionally, we employ GPT-4o \cite{chatgpt} to evaluate the generated solutions based on the original images and questions, filtering out duplicates and instances with inconsistent correctness labels. This process results in 38K examples for post-training on Qwen2.5-VL-7B but achieving an accuracy of 70.3\% on MathVista, which is comparable to the performance of Qwen2.5-VL-DP. These results demonstrate both the quality assurance of our generated data and its effectiveness in enhancing MLLM's ability.

\subsection{Generalizability of Qwen-VL-DP}
The proposed Qwen-VL-DP model exhibits strong performance in multimodal mathematical reasoning tasks. To further evaluate its generalization capabilities, we conduct experiments on the MMMU benchmark \cite{yue2023mmmu}, which spans a wide range of disciplines and domains. As shown in Table~\ref{tab:mmmubaseline}, the Qwen-VL-DP model, post-trained on MathV-DP, consistently outperforms the base model as well as several other open-source MLLMs. These results highlight the model’s ability to generalize effectively to diverse downstream multimodal understanding and reasoning tasks. Notably, the post-training with our synthetic dataset not only preserves but also enhances the model's reasoning performance in other domains, demonstrating the robustness and generalizability of Qwen-VL-DP.

\begin{table}[h]\small
\centering
\caption{Comparison on the MMMU benchmark.}
\renewcommand{\arraystretch}{1.33}
\setlength{\tabcolsep}{5.0mm}{
\begin{tabular}{c|c}
\hline
			
\multicolumn{1}{c|}{\multirow{1}{*}{\textbf{Model}}} & \textbf{MMMU} \\ \hline 

\multicolumn{1}{c|}{Random Chance} & 22.1 \\
\multicolumn{1}{c|}{Frequent Guess} & 26.8 \\
\multicolumn{1}{c|}{SPHINX-13B} & 32.9   \\
\multicolumn{1}{c|}{InstructBLIP-7B} & 32.9   \\
\multicolumn{1}{c|}{LLaVA-1.5-13B} & 36.4   \\ \hline
\multicolumn{1}{c|}{Qwen2-VL-7B} & 47.8 \\
\multicolumn{1}{c|}{Qwen2-VL-DP} & 49.4 \\ 
\multicolumn{1}{c|}{Qwen2.5-VL-7B} & 58.6 \\ 
\multicolumn{1}{c|}{Qwen2.5-VL-DP} & \textbf{59.4} \\ \hline
\end{tabular}}

\label{tab:mmmubaseline}
\end{table}

\section{Conclusions and Future Work}
In this work, we proposed MathV-DP, a novel dataset that enriched multimodal mathematical reasoning with diverse solving perspectives and reflective supervision. 
Building upon Qwen-VL
, we introduced Qwen-VL-DP, trained via both supervised fine-tuning 
and group relative policy optimization, 
a rule-based reinforcement learning method tailored to reward correctness, diversity, and discrimination of multiple solutions. 
Our experiments on MathVista
and Math-V benchmarks demonstrated that incorporating diverse reasoning perspectives significantly enhanced both the accuracy and generative diversity of MLLMs. 
These findings highlight the importance of moving beyond one-to-one image-text supervision, advocating for a shift towards learning from multiple valid solving perspectives.

While our framework improves both reasoning accuracy and response diversity, the diversity learned from MathV-DP remains implicit. The model can generate multiple valid reasoning trajectories and solution perspectives for a given problem, but it cannot explicitly control which perspective is adopted in a particular response during inference. 
Enabling controllable reasoning generation would allow users to specify desired reasoning perspectives or solution strategies, thereby enabling MLLMs to produce corresponding reasoning trajectories in a more interpretable, flexible, and controllable manner. Such capability may further enhance the practical usability of multimodal mathematical reasoning systems.

\section*{Acknowledgments}
This research is supported by A*STAR, CISCO Systems (USA) Pte. Ltd and National University of Singapore under its Cisco-NUS Accelerated Digital Economy Corporate Laboratory (Award I21001E0002). This work is also supported by the Central Guidance on Local Science and Technology Development Fund of Shanghai City (No.YDZX20253100002004), and Fundamental and Interdisciplinary Disciplines Breakthrough Plan of the Ministry of Education of China (No. JYB2025XDXM116).



 
\bibliography{IEEEabrv,reference}

\begin{thebibliography}{10}
\providecommand{\url}[1]{#1}
\csname url@samestyle\endcsname
\providecommand{\newblock}{\relax}
\providecommand{\bibinfo}[2]{#2}
\providecommand{\BIBentrySTDinterwordspacing}{\spaceskip=0pt\relax}
\providecommand{\BIBentryALTinterwordstretchfactor}{4}
\providecommand{\BIBentryALTinterwordspacing}{\spaceskip=\fontdimen2\font plus
\BIBentryALTinterwordstretchfactor\fontdimen3\font minus \fontdimen4\font\relax}
\providecommand{\BIBforeignlanguage}[2]{{%
\expandafter\ifx\csname l@#1\endcsname\relax
\typeout{** WARNING: IEEEtran.bst: No hyphenation pattern has been}%
\typeout{** loaded for the language `#1'. Using the pattern for}%
\typeout{** the default language instead.}%
\else
\language=\csname l@#1\endcsname
\fi
#2}}
\providecommand{\BIBdecl}{\relax}
\BIBdecl

\bibitem{wang2023selfconsistency}
X.~Wang, J.~Wei, D.~Schuurmans, Q.~V. Le, E.~H. Chi, S.~Narang, A.~Chowdhery, and D.~Zhou, ``Self-consistency improves chain of thought reasoning in language models,'' in \emph{The Eleventh International Conference on Learning Representations}, 2023.

\bibitem{zhou2023least-to-most}
D.~Zhou \emph{et~al.}, ``Least-to-most prompting enables complex reasoning in large language models,'' in \emph{The Eleventh International Conference on Learning Representations}, 2023.

\bibitem{huang2026step}
A.~Huang \emph{et~al.}, ``Step 3.5 flash: Open frontier-level intelligence with 11b active parameters,'' \emph{arXiv preprint arXiv:2602.10604}, 2026.

\bibitem{luo2023wizard}
H.~Luo, Q.~Sun, C.~Xu, P.~Zhao, J.~Lou, C.~Tao, X.~Geng, Q.~Lin, S.~Chen, and D.~Zhang, ``Wizardmath: Empowering mathematical reasoning for large language models via reinforced evol-instruct,'' \emph{CoRR}, vol. abs/2308.09583, 2023.

\bibitem{yue2023mammoth}
X.~Yue, X.~Qu, G.~Zhang, Y.~Fu, W.~Huang, H.~Sun, Y.~Su, and W.~Chen, ``Mammoth: Building math generalist models through hybrid instruction tuning,'' \emph{CoRR}, vol. abs/2309.05653, 2023.

\bibitem{gou2023tora}
Z.~Gou, Z.~Shao, Y.~Gong, Y.~Shen, Y.~Yang, M.~Huang, N.~Duan, and W.~Chen, ``Tora: {A} tool-integrated reasoning agent for mathematical problem solving,'' \emph{CoRR}, vol. abs/2309.17452, 2023.

\bibitem{jiang2023formaltheorem}
A.~Q. Jiang, S.~Welleck, J.~P. Zhou, T.~Lacroix, J.~Liu, W.~Li, M.~Jamnik, G.~Lample, and Y.~Wu, ``Draft, sketch, and prove: Guiding formal theorem provers with informal proofs,'' in \emph{The Eleventh International Conference on Learning Representations}, 2023.

\bibitem{lu2023mathvista}
P.~Lu, H.~Bansal, T.~Xia, J.~Liu, C.~Li, H.~Hajishirzi, H.~Cheng, K.~Chang, M.~Galley, and J.~Gao, ``Mathvista: Evaluating math reasoning in visual contexts with gpt-4v, bard, and other large multimodal models,'' \emph{CoRR}, vol. abs/2310.02255, 2023.

\bibitem{liu2023llava}
H.~Liu, C.~Li, Q.~Wu, and Y.~J. Lee, ``Visual instruction tuning,'' in \emph{Advances in Neural Information Processing Systems}, A.~Oh, T.~Naumann, A.~Globerson, K.~Saenko, M.~Hardt, and S.~Levine, Eds., 2023.

\bibitem{bai2023qwenvl}
J.~Bai, S.~Bai, S.~Yang, S.~Wang, S.~Tan, P.~Wang, J.~Lin, C.~Zhou, and J.~Zhou, ``Qwen-vl: {A} frontier large vision-language model with versatile abilities,'' \emph{CoRR}, vol. abs/2308.12966, 2023.

\bibitem{guo2023unkvqa}
Y.~Guo, F.~Jiao, Z.~Shen, L.~Nie, and M.~S. Kankanhalli, ``{UNK-VQA:} {A} dataset and {A} probe into multi-modal large models' abstention ability,'' \emph{CoRR}, vol. abs/2310.10942, 2023.

\bibitem{gpt4v}
OpenAI-b, ``Gpt-4v,'' \url{https://openai.com/research/gpt-4v-system-card}.

\bibitem{gemini}
Google, ``Gemini,'' \url{https://gemini.google.com}.

\bibitem{ericsson1980verbal}
K.~A. Ericsson and H.~A. Simon, ``Verbal reports as data.'' \emph{Psychological review}, vol.~87, no.~3, p. 215, 1980.

\bibitem{jason2022cotreason}
J.~Wei, X.~Wang, D.~Schuurmans, M.~Bosma, B.~Ichter, F.~Xia, E.~H. Chi, Q.~V. Le, and D.~Zhou, ``Chain-of-thought prompting elicits reasoning in large language models,'' in \emph{Advances in Neural Information Processing Systems}, 2022.

\bibitem{hsieh2023distilling}
C.-Y. Hsieh, C.-L. Li, C.-K. Yeh, H.~Nakhost, Y.~Fujii, A.~Ratner, R.~Krishna, C.-Y. Lee, and T.~Pfister, ``Distilling step-by-step! outperforming larger language models with less training data and smaller model sizes,'' \emph{arXiv preprint arXiv:2305.02301}, 2023.

\bibitem{OpenAIo1}
OpenAI-c, ``Introducing openai o1,'' \url{https://openai.com/o1/}.

\bibitem{dai2024instructblip}
W.~Dai, J.~Li, D.~Li, A.~M.~H. Tiong, J.~Zhao, W.~Wang, B.~Li, P.~N. Fung, and S.~Hoi, ``Instructblip: Towards general-purpose vision-language models with instruction tuning,'' \emph{Advances in Neural Information Processing Systems}, vol.~36, 2024.

\bibitem{sutton1998reinforcement}
R.~S. Sutton, A.~G. Barto \emph{et~al.}, \emph{Reinforcement learning: An introduction}.\hskip 1em plus 0.5em minus 0.4em\relax MIT press Cambridge, 1998, vol.~1, no.~1.

\bibitem{guo2025deepseek}
D.~Guo \emph{et~al.}, ``Deepseek-r1: Incentivizing reasoning capability in llms via reinforcement learning,'' \emph{arXiv preprint arXiv:2501.12948}, 2025.

\bibitem{radford2019language}
A.~Radford, J.~Wu, R.~Child, D.~Luan, D.~Amodei, I.~Sutskever \emph{et~al.}, ``Language models are unsupervised multitask learners,'' \emph{OpenAI blog}, vol.~1, no.~8, p.~9, 2019.

\bibitem{chen2025r1v}
L.~Chen, L.~Li, H.~Zhao, Y.~Song, and Vinci, ``R1-v: Reinforcing super generalization ability in vision-language models with less than \$3,'' \url{https://github.com/Deep-Agent/R1-V}, 2025, accessed: 2025-02-02.

\bibitem{yang2025r1}
Y.~Yang \emph{et~al.}, ``R1-onevision: Advancing generalized multimodal reasoning through cross-modal formalization,'' \emph{arXiv preprint arXiv:2503.10615}, 2025.

\bibitem{Qwen2.5-VL}
S.~Bai \emph{et~al.}, ``Qwen2.5-vl technical report,'' \emph{arXiv preprint arXiv:2502.13923}, 2025.

\bibitem{Qwen2-VL}
P.~Wang \emph{et~al.}, ``Qwen2-vl: Enhancing vision-language model's perception of the world at any resolution,'' \emph{arXiv preprint arXiv:2409.12191}, 2024.

\bibitem{shao2024deepseekmath}
Z.~Shao \emph{et~al.}, ``Deepseekmath: Pushing the limits of mathematical reasoning in open language models,'' \emph{arXiv preprint arXiv:2402.03300}, 2024.

\bibitem{wang2024mathv}
K.~Wang, J.~Pan, W.~Shi, Z.~Lu, M.~Zhan, and H.~Li, ``Measuring multimodal mathematical reasoning with math-vision dataset,'' \emph{arXiv preprint arXiv:2402.14804}, 2024.

\bibitem{radford2021clip}
A.~Radford \emph{et~al.}, ``Learning transferable visual models from natural language supervision,'' in \emph{International conference on machine learning}, 2021, pp. 8748--8763.

\bibitem{li2022blip}
J.~Li, D.~Li, C.~Xiong, and S.~Hoi, ``Blip: Bootstrapping language-image pre-training for unified vision-language understanding and generation,'' in \emph{International conference on machine learning}, 2022, pp. 12\,888--12\,900.

\bibitem{li2024mm}
H.~Li, Z.~Yang, Y.~Ma, Y.~Bin, Y.~Yang, and T.-S. Chua, ``Mm-forecast: A multimodal approach to temporal event forecasting with large language models,'' in \emph{ACM Multimedia 2024}, 2024.

\bibitem{bin2024gallerygpt}
Y.~Bin, W.~Shi, Y.~Ding, Z.~Hu, Z.~Wang, Y.~Yang, S.-K. Ng, and H.~T. Shen, ``Gallerygpt: Analyzing paintings with large multimodal models,'' in \emph{ACM Multimedia 2024}, 2024.

\bibitem{zhu2023minigpt4}
D.~Zhu, J.~Chen, X.~Shen, X.~Li, and M.~Elhoseiny, ``Minigpt-4: Enhancing vision-language understanding with advanced large language models,'' \emph{CoRR}, vol. abs/2304.10592, 2023.

\bibitem{ye2023mplug}
Q.~Ye \emph{et~al.}, ``mplug-owl: Modularization empowers large language models with multimodality,'' \emph{arXiv preprint arXiv:2304.14178}, 2023.

\bibitem{lin2023sphinx}
Z.~Lin \emph{et~al.}, ``Sphinx: The joint mixing of weights, tasks, and visual embeddings for multi-modal large language models,'' \emph{arXiv preprint arXiv:2311.07575}, 2023.

\bibitem{hu2024minicpm}
S.~Hu \emph{et~al.}, ``Minicpm: Unveiling the potential of small language models with scalable training strategies,'' \emph{arXiv preprint arXiv:2404.06395}, 2024.

\bibitem{huang2025vision}
W.~Huang, B.~Jia, Z.~Zhai, S.~Cao, Z.~Ye, F.~Zhao, Z.~Xu, X.~Tang, Y.~Hu, and S.~Lin, ``Vision-r1: Incentivizing reasoning capability in multimodal large language models,'' \emph{arXiv preprint arXiv:2503.06749}, 2025.

\bibitem{wang2025adapting}
H.~Wang, C.~Lai, and W.~Ge, ``Adapting multimodal large language models for video question answering by capturing question-critical and coherent moments,'' \emph{IEEE Transactions on Multimedia}, 2025.

\bibitem{xu2026mmtot}
N.~Xu, Z.~Lu, H.~Tian, B.~Zheng, J.~Cao, and A.-A. Liu, ``Mmtot: Multi-modal token-of-thought reasoning for large models,'' \emph{IEEE Transactions on Multimedia}, 2026.

\bibitem{zhang2023multicot}
Z.~Zhang, A.~Zhang, M.~Li, H.~Zhao, G.~Karypis, and A.~Smola, ``Multimodal chain-of-thought reasoning in language models,'' \emph{arXiv preprint arXiv:2302.00923}, 2023.

\bibitem{wang2024tsciq}
L.~Wang, Y.~Hu, J.~He, X.~Xu, N.~Liu, H.~Liu, and H.~T. Shen, ``T-sciq: Teaching multimodal chain-of-thought reasoning via large language model signals for science question answering,'' in \emph{Proceedings of the AAAI Conference on Artificial Intelligence}, vol.~38, 2024, pp. 19\,162--19\,170.

\bibitem{chen2023chain}
F.~Chen and Y.~Feng, ``Chain-of-thought prompt distillation for multimodal named entity and multimodal relation extraction,'' \emph{arXiv preprint arXiv:2306.14122}, 2023.

\bibitem{li2024-multimodal-arxiv}
L.~Li, Y.~Wang, R.~Xu, P.~Wang, X.~Feng, L.~Kong, and Q.~Liu, ``Multimodal {A}r{X}iv: A dataset for improving scientific comprehension of large vision-language models,'' in \emph{Proceedings of the 62nd Annual Meeting of the Association for Computational Linguistics}, 2024, pp. 14\,369--14\,387.

\bibitem{hu2023vpd}
Y.~Hu, O.~Stretcu, C.~Lu, K.~Viswanathan, K.~Hata, E.~Luo, R.~Krishna, and A.~Fuxman, ``Visual program distillation: Distilling tools and programmatic reasoning into vision-language models,'' \emph{CoRR}, vol. abs/2312.03052, 2023.

\bibitem{zheng2023ddcot}
G.~Zheng, B.~Yang, J.~Tang, H.-Y. Zhou, and S.~Yang, ``Ddcot: Duty-distinct chain-of-thought prompting for multimodal reasoning in language models,'' \emph{Advances in Neural Information Processing Systems}, vol.~36, pp. 5168--5191, 2023.

\bibitem{mathllava}
W.~Shi, Z.~Hu, Y.~Bin, J.~Liu, Y.~Yang, S.-K. Ng, L.~Bing, and R.~K.-W. Lee, ``Math-llava: Bootstrapping mathematical reasoning for multimodal large language models,'' \emph{arXiv preprint arXiv:2406.17294}, 2024.

\bibitem{yue2023mmmu}
X.~Yue \emph{et~al.}, ``{MMMU:} {A} massive multi-discipline multimodal understanding and reasoning benchmark for expert {AGI},'' \emph{CoRR}, vol. abs/2311.16502, 2023.

\bibitem{littman1996reinforcement}
M.~Littman and A.~Moore, ``Reinforcement learning: A survey, journal of artificial intelligence research 4,'' 1996.

\bibitem{brown2020language}
T.~Brown \emph{et~al.}, ``Language models are few-shot learners,'' \emph{Advances in neural information processing systems}, vol.~33, pp. 1877--1901, 2020.

\bibitem{bai2022training}
Y.~Bai \emph{et~al.}, ``Training a helpful and harmless assistant with reinforcement learning from human feedback,'' \emph{arXiv preprint arXiv:2204.05862}, 2022.

\bibitem{schulman2017ppo}
J.~Schulman, F.~Wolski, P.~Dhariwal, A.~Radford, and O.~Klimov, ``Proximal policy optimization algorithms,'' \emph{arXiv preprint arXiv:1707.06347}, 2017.

\bibitem{rafailov2023dpo}
R.~Rafailov, A.~Sharma, E.~Mitchell, C.~D. Manning, S.~Ermon, and C.~Finn, ``Direct preference optimization: Your language model is secretly a reward model,'' \emph{Advances in Neural Information Processing Systems}, vol.~36, pp. 53\,728--53\,741, 2023.

\bibitem{team2025kimi}
K.~Team \emph{et~al.}, ``Kimi k1. 5: Scaling reinforcement learning with llms,'' \emph{arXiv preprint arXiv:2501.12599}, 2025.

\bibitem{luo2025ursa}
R.~Luo, Z.~Zheng, Y.~Wang, Y.~Yu, X.~Ni, Z.~Lin, J.~Zeng, and Y.~Yang, ``Ursa: Understanding and verifying chain-of-thought reasoning in multimodal mathematics,'' \emph{arXiv e-prints}, pp. arXiv--2501, 2025.

\bibitem{yuan2025more}
X.~Yuan, Y.~Ding, Y.~Bin, W.~Shao, J.~Cai, J.~Song, Y.~Yang, and H.~T. Shen, ``More than one teacher: Adaptive multi-guidance policy optimization for diverse exploration,'' \emph{arXiv preprint arXiv:2510.02227}, 2025.

\bibitem{fan2025sophiavl}
K.~Fan, K.~Feng, H.~Lyu, D.~Zhou, and X.~Yue, ``Sophiavl-r1: Reinforcing mllms reasoning with thinking reward,'' \emph{arXiv preprint arXiv:2505.17018}, 2025.

\bibitem{sharif2026sight}
O.~Sharif, E.~Hossain, and P.~Ng, ``From sight to insight: Improving visual reasoning capabilities of multimodal models via reinforcement learning,'' \emph{arXiv preprint arXiv:2601.00215}, 2026.

\bibitem{zhang2024rest}
D.~Zhang, S.~Zhoubian, Z.~Hu, Y.~Yue, Y.~Dong, and J.~Tang, ``Rest-mcts*: Llm self-training via process reward guided tree search,'' \emph{Advances in Neural Information Processing Systems}, vol.~37, pp. 64\,735--64\,772, 2024.

\bibitem{peng2024multimath}
S.~Peng, D.~Fu, L.~Gao, X.~Zhong, H.~Fu, and Z.~Tang, ``Multimath: Bridging visual and mathematical reasoning for large language models,'' \emph{arXiv preprint arXiv:2409.00147}, 2024.

\bibitem{gao2023gllava}
J.~Gao \emph{et~al.}, ``G-llava: Solving geometric problem with multi-modal large language model,'' \emph{arXiv preprint arXiv:2312.11370}, 2023.

\bibitem{team2023gemini}
G.~Team \emph{et~al.}, ``Gemini: a family of highly capable multimodal models,'' \emph{arXiv preprint arXiv:2312.11805}, 2023.

\bibitem{anthropic2024claude}
A.~Anthropic, ``The claude 3 model family: Opus, sonnet, haiku,'' \emph{Claude-3 Model Card}, 2024.

\bibitem{chatgpt}
OpenAI-a, ``Chatgpt,'' \url{https://chat.openai.com}.

\bibitem{llama4}
Meta-AI, ``The llama 4 herd: The beginning of a new era of natively multimodal ai innovation,'' \url{https://ai.meta.com/blog/llama-4-multimodal-intelligence/}.

\bibitem{laurenccon2024obelics}
H.~Lauren{\c{c}}on \emph{et~al.}, ``Obelics: An open web-scale filtered dataset of interleaved image-text documents,'' \emph{Advances in Neural Information Processing Systems}, vol.~36, 2024.

\bibitem{gao2023llama}
P.~Gao \emph{et~al.}, ``Llama-adapter v2: Parameter-efficient visual instruction model,'' \emph{arXiv preprint arXiv:2304.15010}, 2023.

\bibitem{zhang2023llavar}
Y.~Zhang, R.~Zhang, J.~Gu, Y.~Zhou, N.~Lipka, D.~Yang, and T.~Sun, ``Llavar: Enhanced visual instruction tuning for text-rich image understanding,'' \emph{arXiv preprint arXiv:2306.17107}, 2023.

\bibitem{liu2024llava-1.5}
H.~Liu, C.~Li, Y.~Li, and Y.~J. Lee, ``Improved baselines with visual instruction tuning,'' in \emph{Proceedings of the IEEE/CVF Conference on Computer Vision and Pattern Recognition}, 2024, pp. 26\,296--26\,306.

\bibitem{omnilmm}
OpenBMB, ``Large multi-modal models for strong performance and efficient deployment.'' 2024, \url{https://github.com/OpenBMB/OmniLMM}.

\bibitem{liu2024llavanext}
\BIBentryALTinterwordspacing
H.~Liu, C.~Li, Y.~Li, B.~Li, Y.~Zhang, S.~Shen, and Y.~J. Lee, ``Llava-next: Improved reasoning, ocr, and world knowledge,'' 2024. [Online]. Available: \url{https://llava-vl.github.io/blog/2024-01-30-llava-next/}
\BIBentrySTDinterwordspacing

\bibitem{mathpuma}
W.~Zhuang, X.~Huang, X.~Zhang, and J.~Zeng, ``Math-puma: Progressive upward multimodal alignment to enhance mathematical reasoning,'' in \emph{Proceedings of the AAAI Conference on Artificial Intelligence}, vol.~39, no.~24, 2025, pp. 26\,183--26\,191.

\bibitem{yao2024mulberry}
H.~Yao \emph{et~al.}, ``Mulberry: Empowering mllm with o1-like reasoning and reflection via collective monte carlo tree search,'' \emph{arXiv preprint arXiv:2412.18319}, 2024.

\bibitem{shypula2025evaluating}
A.~Shypula, S.~Li, B.~Zhang, V.~Padmakumar, K.~Yin, and O.~Bastani, ``Evaluating the diversity and quality of llm generated content,'' \emph{arXiv preprint arXiv:2504.12522}, 2025.

\bibitem{reimers-2019-sentence-bert}
N.~Reimers and I.~Gurevych, ``Sentence-bert: Sentence embeddings using siamese bert-networks,'' in \emph{Proceedings of the 2019 Conference on Empirical Methods in Natural Language Processing}.\hskip 1em plus 0.5em minus 0.4em\relax Association for Computational Linguistics, 11 2019.

\bibitem{long2024llms}
L.~Long, R.~Wang, R.~Xiao, J.~Zhao, X.~Ding, G.~Chen, and H.~Wang, ``On llms-driven synthetic data generation, curation, and evaluation: A survey,'' \emph{arXiv preprint arXiv:2406.15126}, 2024.

\end{thebibliography}
\bibliographystyle{IEEEtran}
%













\section{Biography Section}
\begin{IEEEbiographynophoto}{Wenhao Shi}
received the M.Sc. degree in computer science and B.Sc. degree in mathematics from University of Electronic Science and Technology of China, Chengdu, China. His research interests include natural language processing, vision-language and multimodal large language models.
\end{IEEEbiographynophoto}

\begin{IEEEbiographynophoto}{Zhiqiang Hu}
received Ph.D. degree with the Singapore University of Technology and Design, Singapore. His research interests include text style transfer, large language models, and mathematical reasoning. He is currently with Tencent, Beijing, China. He received the M.Sc. degree in Computer Science from the University of Electronic Science and Technology of China, Chengdu, China, and the B.Sc. degree in applied mathematics from UESTC.
\end{IEEEbiographynophoto}

\begin{IEEEbiographynophoto}{Yi Bin}
is with School of Computer 
Science and Technology, Tongji University, Shanghai, China. He received the Ph.D. degree from the University of Electronic Science and Technology of China, Chengdu, China, in 2020. His research interests include multimedia understanding, natural language processing, multimodal large language models, and machine learning.
\end{IEEEbiographynophoto}

\begin{IEEEbiographynophoto}{Guoqing Wang}
received the B.Sc. and M.Sc. degrees from the China University of Mining and Technology, Xuzhou, China, in 2014 and 2017, respectively, and the Ph.D. degree from the University of New South Wales, Sydney, NSW, Australia, in 2021. He is currently an Associate Professor with the University of Electronic Science and Technology of China, Chengdu, China. His research interests include computer vision, deep learning, visual understanding, transfer learning, and intelligent unmanned systems.
\end{IEEEbiographynophoto}

\begin{IEEEbiographynophoto}{Xing Xu}
received the B.E. and M.E. degrees from Huazhong University of Science and Technology, China, in 2009 and 2012, respectively, and the Ph.D. degree from Kyushu University, Japan, in 2015. He is currently a Professor with Tongji University, Shanghai, China. His research interests mainly focus on multimedia information retrieval, especially cross-modal retrieval and knowledge transfer in integrating language and vision.
\end{IEEEbiographynophoto}

\begin{IEEEbiographynophoto}{Yang Yang (Senior Member)}
received the B.Sc. degree from
Jilin University, Changchun, China, in 2006, the M.Sc. degree from Peking
University, Beijing, China, in 2009, and the Ph.D. degree from The University
of Queensland, Brisbane, QLD, Australia, in 2012, all in computer science.
He is currently with the University of Electronic Science and Technology
of China, Chengdu, China. His current research interests include multimedia
content analysis, computer vision, and social media analytics.
\end{IEEEbiographynophoto}

\begin{IEEEbiographynophoto}{See-Kiong Ng (Senior Member)}
received the B.Sc., M.Sc., and Ph.D. degrees in computer science from Carnegie Mellon University, Pittsburgh, PA, USA, in 1989, 1994, and 1998, respectively, and the M.Sc. degree in Artificial Intelligence from the University of Pennsylvania, Philadelphia, PA, USA, in 1991. He is currently a Professor of Practice with the Department of Computer Science, School of Computing, National University of Singapore, Singapore, and the Director of Translational Research at the Institute of Data Science, NUS. His research interests include artificial intelligence, data science, machine learning, multimodal intelligence, bioinformatics, and smart city analytics.
\end{IEEEbiographynophoto}

 




\vfill

\end{document}